# Decision-Theoretic Bidding Based on Learned Density Models in Simultaneous, Interacting Auctions

**Peter Stone**                                                    PSTONE@CS.UTEXAS.EDU
*Dept. of Computer Sciences, The University of Texas at Austin*
*1 University Station C0500, Austin, Texas 78712-1188 USA*

**Robert E. Schapire**                                          SCHAPIRE@CS.PRINCETON.EDU
*Department of Computer Science, Princeton University*
*35 Olden Street, Princeton, NJ 08544 USA*

**Michael L. Littman**                                          MLITTMAN@CS.RUTGERS.EDU
*Dept. of Computer Science, Rutgers University*
*Piscataway, NJ 08854-8019 USA*

**János A. Csirik**                                              JANOS@POBOX.COM
*D. E. Shaw & Co.*
*120 W 45th St, New York, NY 10036 USA*

**David McAllester**                                            MCALLESTER@TTI-CHICAGO.EDU
*Toyota Technological Institute at Chicago*
*1427 East 60th Street, Chicago IL, 60637 USA*

## Abstract

Auctions are becoming an increasingly popular method for transacting business, especially over the Internet. This article presents a general approach to building autonomous bidding agents to bid in multiple simultaneous auctions for interacting goods. A core component of our approach learns a model of the empirical price dynamics based on past data and uses the model to analytically calculate, to the greatest extent possible, optimal bids. We introduce a new and general boosting-based algorithm for conditional density estimation problems of this kind, *i.e.*, supervised learning problems in which the goal is to estimate the entire conditional distribution of the real-valued label. This approach is fully implemented as ATTac-2001, a top-scoring agent in the second Trading Agent Competition (TAC-01). We present experiments demonstrating the effectiveness of our boosting-based price predictor relative to several reasonable alternatives.

## 1. Introduction

Auctions are an increasingly popular method for transacting business, especially over the Internet. In an auction for a single good, it is straightforward to create automated bidding strategies—an agent could keep bidding until reaching a target reserve price, or it could monitor the auction and place a winning bid just before the closing time (known as *sniping*).

When bidding for multiple interacting goods in simultaneous auctions, on the other hand, agents must be able to reason about uncertainty and make complex value assessments. For example, an agent bidding on one's behalf in separate auctions for a camera and flash may end up buying the flash and then not being able to find an affordable camera. Alternatively, if bidding for the same good in several auctions, it may purchase two flashes when only one was needed.

This article makes three main contributions. The first contribution is a general approach to building autonomous bidding agents to bid in multiple simultaneous auctions for interacting goods. We start with the observation that the key challenge in auctions is the prediction of eventual prices of goods: with complete knowledge of eventual prices, there are direct methods for determining the optimal bids to place. Our guiding principle is to have the agent model its *uncertainty* in eventual prices and, to the greatest extent possible, analytically calculate optimal bids.

To attack the price prediction problem, we propose a machine-learning approach: gather examples of previous auctions and the prices paid in them, then use machine-learning methods to predict these prices based on available features in the auction. Moreover, for our strategy, we needed to be able to model the uncertainty associated with predicted prices; in other words, we needed to be able to sample from a predicted distribution of prices given the current state of the game. This can be viewed as a *conditional density estimation* problem, that is, a supervised learning problem in which the goal is to estimate the entire distribution of a real-valued label given a description of current conditions, typically in the form of a feature vector. The second main contribution of this article is a new algorithm for solving such general problems based on boosting (Freund & Schapire, 1997; Schapire & Singer, 1999).

The third contribution of this article is a complete description of a prototype implementation of our approach in the form of ATTac-2001, a top-scoring agent[1] in the second Trading Agent Competition (TAC-01) that was held in Tampa Bay, FL on October 14, 2001 (Wellman, Greenwald, Stone, & Wurman, 2003a). The TAC domain was the main motivation for the innovations reported here. ATTac-2001 builds on top of ATTac-2000 (Stone, Littman, Singh, & Kearns, 2001), the top-scoring agent at TAC-00, but introduces a fundamentally new approach to creating autonomous bidding agents.

We present details of ATTac-2001 as an instantiation of its underlying principles that we believe have applications in a wide variety of bidding situations. ATTac-2001 uses a predictive, data-driven approach to bidding based on expected marginal values of all available goods. In this article, we present empirical results demonstrating the robustness and effectiveness of ATTac-2001's adaptive strategy. We also report on ATTac-2001's performance at TAC-01 and TAC-02 and reflect on some of the key issues raised during the competitions.

The remainder of the article is organized as follows. In Section 2, we present our general approach to bidding for multiple interacting goods in simultaneous auctions. In Section 3, we summarize TAC, the substrate domain for our work. Section 4 describes our boosting-based price predictor. In Section 5, we give the details of ATTac-2001. In Section 6, we present empirical results including a summary of ATTac-2001's performance in TAC-01, controlled experiments isolating the successful aspects of ATTac-2001, and controlled experiments illustrating some of the lessons learned during the competition. A discussion and summary of related work is provided in Sections 7 and 8.

## 2. General Approach

In a wide variety of decision-theoretic settings, it is useful to be able to evaluate hypothetical situations. In computer chess, for example, a static board evaluator is used to heuristically measure which player is ahead and by how much in a given board situation. The scenario is similar in auction domains, and our bidding agent ATTac-2001 uses a situation evaluator, analogous to the static board evaluator, which estimates the agent's expected profit in a hypothetical future situation. This "profit predictor" has a wide variety of uses in the agent. For example, to determine the value of an item, the agent compares the predicted profit assuming the item is already owned to the predicted profit assuming that the item is not available.

---

1. Top-scoring by one metric, and second place by another.

Given prices for goods, one can often compute a set of purchases and an allocation that maximizes profit.[2] Similarly, if closing prices are known, they can be treated as fixed, and optimal bids can be computed (bid high for anything you want to buy). So, one natural profit predictor is simply to calculate the profit of optimal purchases under fixed predicted prices. (The predicted prices can, of course, be different in different situations, e.g., previous closing prices can be relevant to predicting future closing prices.)

A more sophisticated approach to profit prediction is to construct a model of the probability distribution over possible future prices and to place bids that maximize *expected* profit. An approximate solution to this difficult optimization problem can be created by stochastically sampling possible prices and computing a profit prediction as above for each sampled price. A sampling-based scheme for profit prediction is important for modeling uncertainty and the value of gaining information, *i.e.*, reducing the price uncertainty.

Section 2.1 formalizes this latter approach within a simplified sequential auction model. This abstraction illustrates some of the decision-making issues in our full sampling-based approach presented in Section 2.2. The full setting that our approach addresses is considerably more complex than the abstract model, but our simplifying assumptions allow us to focus on a core challenge of the full scenario. Our guiding principle is to make decision-theoretically optimal decisions given profit predictions for hypothetical future situations.[3]

## 2.1 Simplified Abstraction

In the simple model, there are $n$ items to be auctioned off in sequence (first item 0, then item 1, etc.). The bidder must place a bid $r_i$ for each item $i$, and after each bid, a closing price $y_i$ is chosen for the corresponding item from a distribution specific to the item. If the bid matches or exceeds the closing price, $r_i \geq y_i$, the bidder holds item $i$, $h_i = 1$. Otherwise, the bidder does not hold the item, $h_i = 0$. The bidder's utility $v(H)$ is a function of its final vector of holdings $H = (h_0, \ldots, h_{n-1})$ and its cost is a function of the holdings and the vector of closing prices, $H \cdot Y$. We will formalize the problem of optimal bid selection and develop a series of approximations to make the problem solvable.

---

2. The problem is computationally difficult in general, but has been solved effectively in the non-trivial TAC setting (Greenwald & Boyan, 2001; Stone et al., 2001).

3. An alternative approach would be to abstractly calculate the Bayes-Nash equilibrium (Harsanyi, 1968) for the game and play the optimal strategy. We dismissed this approach because of its intractability in realistically complex situations, including TAC. Furthermore, even if we were able to approximate the equilibrium strategy, it is reasonable to assume that our opponents would not play optimal strategies. Thus, we could gain additional advantage by tuning our approach to our opponents' *actual* behavior as observed in the earlier rounds, which is essentially the strategy we adopted.

### 2.1.1 Exact Value

What is the *value* of the auction, that is, the bidder's expected profit (utility minus cost) for bidding optimally for the rest of the auction? If a bidder knows this value, it can make its next bid to be one that maximizes its expected profit. The value is a function of the bidder's current holdings $H$ and the current item to bid on, $i$. It can be expressed as

$$\text{value}(i, H) = \max_{r_i} E_{y_i} \max_{r_{i+1}} E_{y_{i+1}} \dots \max_{r_{n-1}} E_{y_{n-1}} (v(G + H) - G \cdot Y), \tag{1}$$

where the components of $G$ are the new holdings as a result of additional winnings $g_j \equiv r_j \geq y_j$. Note that $H$ only has non-zero entries for items that have already been sold ($\forall j \geq i, H_j = 0$) and $G$ only has non-zero entries for items that have yet to be sold ($\forall j < i, G_j = 0$). Note also that $G$ and $Y$ are fully specified when the $g_j$ and $y_j$ variables (for $j \geq i$) are bound sequentially by the expectation and maximization operators. The idea here is that the bids $r_i$ through $r_{n-1}$ are chosen to maximize value in the context of the possible closing prices $y_j$.

Equation 1 is closely related to the equations defining the value of a finite-horizon partially observable Markov decision process (Papadimitriou & Tsitsiklis, 1987) or a stochastic satisfiability expression (Littman, Majercik, & Pitassi, 2001). Like these other problems, the sequential auction problem is computationally intractable for sufficiently general representations of $v(\cdot)$ (specifically, linear functions of the holdings are not expressive enough to achieve intractability while arbitrary nonlinear functions are).

### 2.1.2 Approximate Value by Reordering

There are three major sources of intractability in Equation 1—the alternation of the maximization and expectation operators (allowing decisions to be conditioned on an exponential number of possible sets of holdings), the large number of maximizations (forcing an exponential number of decisions to be considered), and the large number of expectations (resulting in sums over an exponential number of variable settings).

We attack the problem of interleaved operators by moving all but the first of the maximizations inside the expectations, resulting in an expression that approximates the value:

$$\text{value-est}(i, H) = \max_{r_i} E_{y_i} E_{y_{i+1}} \dots E_{y_{n-1}} \max_{r_{i+1}} \dots \max_{r_{n-1}} (v(G + H) - G \cdot Y). \tag{2}$$

Because the choices for bids $r_{i+1}$ through $r_{n-1}$ appear more deeply nested than the bindings for the closing prices $y_i$ through $y_{n-1}$, they cease to be bids altogether, and instead represent decisions as to whether to purchase goods at given prices. Let $G = \text{opt}(H, i, Y)$ be a vector representing the optimal number of goods to purchase at the prices specified by the vector $Y$ given the current holdings $H$ starting from auction $i$. Conceptually, this can be computed by evaluating

$$\text{opt}(H, i, Y) = \operatorname*{argmax}_{g_i, \dots, g_{n-1}} (v(G + H) - H \cdot Y). \tag{3}$$

Thus, Equation 2 can be written:

$$\text{value-est}(i, H) = \max_{r_i} E_{y_i, \dots, y_{n-1}} (v(\text{opt}(H', i + 1, Y) + H') - \text{opt}(H', i + 1, Y) \cdot Y) \tag{4}$$

where $H'$ is identical to $H$ except the $i$-th component reflects whether item $i$ is won—$r_i \geq y_i$.

Note that there is a further approximation that can be made by computing the expected prices (as point values) before solving the optimization problem. This approach corresponds to further swapping the expectations towards the core of the equation:

$$\text{value-est}(i, H)_{ev} = \max_{r_i}(v(\text{opt}(H', i+1, E_Y) + H') - \text{opt}(H', i+1, E_Y) \cdot E_Y) \quad (5)$$

where $E_Y = E[y_{i+1}, \ldots, y_{n-1}]$, the vector of expected costs of the goods. In the remainder of the article, we refer to methods that use this further approximation from Equation 5 as expected value approaches for reasons that will be come apparent shortly.

The technique of swapping maximization and expectation operators was previously used by Hauskrecht (1997) to generate a bound for solving partially observable Markov decision processes. The decrease of uncertainty when decisions are made makes this approximation an upper bound on the true value of the auction: value-est $\geq$ value. The tightness of the approximations in Equations 2 and 5 depends on the true distributions of the expected prices. For example, if the prices were known in advance with certainty, then both approximations are exact.

### 2.1.3 Approximate Bidding

Given a vector of costs $Y$, the optimization problem $\text{opt}(H, i, Y)$ in Equation 4 is still NP-hard (assuming the representation of the utility function $v(\cdot)$ is sufficiently complex). For many representations of $v(\cdot)$, the optimization problem can be cast as an integer linear program and approximated by using the fractional relaxation instead of the exact optimization problem. This is precisely the approach we have adopted in ATTac (Stone et al., 2001).

### 2.1.4 Approximation via Sampling

Even assuming that $\text{opt}(H, i, Y)$ can be solved in unit time, a literal interpretation of Equation 4 says we'll need to solve this optimization problem for an exponential number of cost vectors (or even more if the probability distributions $\Pr(y_j)$ are continuous). Kearns, Mansour, and Ng (1999) showed that values of partially observable Markov decision processes could be estimated accurately by sampling trajectories instead of exactly computing sums. Littman et al. (2001) did the same for stochastic satisfiability expressions. Applying this idea to Equation 4 leads to the following algorithm.

1. Generate a set $S$ of vectors of closing costs $Y$ according to the product distribution $\Pr(y_i) \times \cdots \times \Pr(y_{n-1})$.

2. For each of these samples, calculate $\text{opt}(H', i+1, Y)$ as defined above and average the results, resulting in the approximation

$$\text{value-est}_s(i, H) = \max_{r_i} \sum_{Y \in S}(v(\text{opt}(H', i+1, Y) + H') - \text{opt}(H', i+1, Y) \cdot Y)/|S|. \quad (6)$$

This expression converges to value-est with increasing sample size.

A remaining challenge in evaluating Equation 6 is computing the real-valued bid $r_i$ that maximizes the value. Note that we want to buy item $i$ precisely at those closing prices for which the value of having the item (minus its cost) exceeds the value of not having the item; this maximizes profit. Thus, to make a positive profit, we are willing to pay up to, but not more than, the difference in value of having the item and not having the item.

Formally, let $H$ be the vector of current holdings and $H_w$ be the holdings modified to reflect winning item $i$. Let $G_w(Y) = \text{opt}(H_w, i+1, Y)$, the optimal set of purchases assuming item $i$ was won, and $G(Y) = \text{opt}(H, i+1, Y)$ the optimal set of purchases assuming otherwise (except in cases of ambiguity, we write simply $G_w$ and $G$ for $G_w(Y)$ and $G(Y)$ respectively). We want to select $r_i$ to achieve the equivalence

$$r_i \geq y_i \equiv \sum_{Y \in S} (v(G_w + H) - G_w \cdot Y)/|S| - y_i \geq \sum_{Y \in S} (v(G + H) - G \cdot Y)/|S|. \qquad (7)$$

Setting

$$r_i = \sum_{Y \in S} ([v(G_w + H) - G_w \cdot Y] - [v(G + H) - G \cdot Y])/|S|. \qquad (8)$$

achieves the equivalence desired in Equation 7, as can be verified by substitution, and therefore bidding the average difference between holding and not holding the item maximizes the value.[4]

## 2.2 The Full Approach

Leveraging from the preceding analysis, we define our sampling-based approach to profit prediction in general simultaneous, multi-unit auctions for interacting goods. In this scenario, let there be $n$ simultaneous, multi-unit auctions for interacting goods $a_0, \ldots, a_{n-1}$. The auctions might close at different times and these times are not, in general, known in advance to the bidders. When an auction closes, let us assume that the $m$ units available are distributed irrevocably to the $m$ highest bidders, who each need to pay the price bid by the $m$th highest bidder. This scenario corresponds to an $m$th price ascending auction.[5] Note that the same bidder may place multiple bids in an auction, and thereby potentially win multiple units. We assume that after the auction closes, the bidders will no longer have any opportunity to acquire additional copies of the goods sold in that auction (*i.e.*, there is no aftermarket).

Our approach is based upon five assumptions. For $G = (g_0, \ldots, g_{n-1}) \in \mathbb{N}^n$, let $v(G) \in \mathbb{R}$ represent the value derived by the agent if it owns $g_i$ units of the commodity being sold in auction $a_i$. Note that $v$ is independent of the *costs* of the commodities. Note further that this representation allows for interacting goods of all kinds, including complementarity and substitutability.[6] The assumptions of our approach are as follows:

1. Closing prices are somewhat, but only somewhat, predictable. That is, given a set of input features $X$, for each auction $a_i$, there exists a sampling rule that outputs a closing price $y_i$ according to a probability distribution of predicted closing prices for $a_i$.

---

4. Note that the strategy for choosing $r_i$ in Equation 8 does not exploit the fact that the sample $S$ contains only a finite set of possibilities for $y_i$, which might make it more robust to inaccuracies in the sampling.
5. For large enough $m$ it is practically the same as the more efficient $m + 1$st auction. We use the $m$th price model because that is what is used in TAC's hotel auctions.
6. Goods are considered complementary if their value as a package is greater than the sum of their individual values; goods are considered substitutable if their value as a package is less than the sum of their individual values.

2. Given a vector of holdings $H = (h_0, \ldots, h_{n-1})$ where $h_i \in \mathbb{N}$ represents the quantity of the commodity being sold in auction $a_i$ that are already owned by the agent, and given a vector of fixed closing prices $Y = (y_0, \ldots, y_{n-1})$, there exists a tractable procedure $\text{opt}(H, Y)$ to determine the optimal set of purchases $(g_0, \ldots, g_{n-1})$ where $g_i \in \mathbb{N}$ represents the number of goods to be purchased in auction $i$ such that

$$v(\text{opt}(H, Y) + H) - \text{opt}(H, Y) \cdot Y \geq v(G + H) - G \cdot Y$$

for all $G \in \mathbb{N}^n$. This procedure corresponds to the optimization problem $\text{opt}(H, i, Y)$ in Equation 3.

3. An individual agent's bids do not have an appreciable effect on the economy (large population assumption).

4. The agent is free to change existing bids in auctions that have not yet closed.

5. Future decisions are made in the presence of complete price information. This assumption corresponds to the operator reordering approximation from the previous section.

While these assumptions are not all true in general, they can be reasonable enough approximations to be the basis for an effective strategy.

By Assumption 3, the price predictor can generate predicted prices prior to considering one's bids. Thus, we can sample from these distributions to produce complete sets of closing prices of all goods.

For each good under consideration, we assume that it is the next one to close. If a different auction closes first, we can then revise our bids later (Assumption 4). Thus, we would like to bid exactly the good's *expected marginal utility* to us. That is, we bid the difference between the expected utilities attainable with and without the good. To compute these expectations, we simply average the utilities of having and not having the good under different price samples as in Equation 8. This strategy rests on Assumption 5 in that we assume that bidding the good's current expected marginal utility cannot adversely affect our future actions, for instance by impacting our future space of possible bids. Note that as time proceeds, the price distributions change in response to the observed price trajectories, thus causing the agent to continually revise its bids.

Table 1 shows pseudo-code for the entire algorithm. A fully detailed description of an instantiation of this approach is given in Section 5.

## 2.3 Example

Consider a camera and a flash with interacting values to an agent as shown in Table 2. Further, consider that the agent estimates that the camera will sell for $40 with probability 25%, $70 with probability 50%, and $95 with probability 25%. Consider the question of what the agent should bid for the flash (in auction $a_0$). The decision pertaining to the camera would be made via a similar analysis.

- Let $H = (h_0, \ldots, h_{n-1})$ be the agent's current holdings in each of the $n$ auctions.

- For $i = 0$ to $n - 1$ (assume auction $i$ is next to close):

  - total-diff = 0

  - counter = 0

  - As time permits:

    * For each auction $a_j, j \neq i$, generate a predicted price sample $y_j$. Let $Y = (y_0, \ldots, y_{i-1}, \infty, y_{i+1}, \ldots, y_{n-1})$.
    * Let $H_w = (h_0, \ldots, h_{i-1}, h_i + 1, h_{i+1}, \ldots, h_{n-1})$, the vector of holdings if the agent wins a unit in auction $a_i$.
    * Compute $G_w = \text{opt}(H_w, Y)$, the optimal set of purchases if the agent wins a unit in auction $a_i$. Note that no additional units of the good will be purchased, since the $i$-th component of $Y$ is $\infty$.
    * Compute $G = \text{opt}(H, Y)$, the optimal set of purchases if the agent never acquires any additional units in the auction $a_i$ and prices are set to $Y$.
    * diff $= [v(G_w + H) - G_w \cdot Y] - [v(G + H) - G \cdot Y]$
    * total-diff = total-diff + diff
    * counter = counter + 1

  - $r$ = total-diff/counter

  - Bid $r$ in auction $a_i$.

Table 1: The decision-theoretic algorithm for bidding in simultaneous, multi-unit, interacting auctions.

|  | *utility* |
|---|---|
| *camera alone* | $50 |
| *flash alone* | 10 |
| *both* | 100 |
| *neither* | 0 |

Table 2: The table of values for all combination of camera and flash in our example.

First, the agent samples from the distribution of possible camera prices. When the price of the camera (sold in auction $a_1$) is $70 in the sample:

- $H = (0, 0), H_w = (1, 0), Y = (\infty, 70)$

- $G_w = \text{opt}(H_w, Y)$ is the best set of purchases the agent can make with the flash, and assuming the camera costs $70. In this case, the only two options are buying the camera or not. Buying the camera yields a profit of $100 - 70 = 30$. Not buying the camera yields a profit of $10 - 0 = 10$. Thus, $G_w = (0, 1)$, and $[v(G_w + H) - G_w \cdot Y] = v(1, 1) - (0, 1) \cdot (\infty, 70) = 100 - 70$.

- Similarly $G = (0, 0)$ (since if the flash is not owned, buying the camera yields a profit of $50 - 70 = -20$, and not buying it yields a profit of $0 - 0 = 0$) and $[v(G + H) - G \cdot Y] = 0$.

- val = 30 − 0 = 30.

Similarly, when the camera is predicted to cost $40, val = 60−10 = 50; and when the camera is predicted to cost $95, val = 10 − 0 = 10. Thus, we expect that 50% of the camera price samples will suggest a flash value of $30, while 25% will lead to a value of $50 and the other 25% will lead to a value of $10. Thus, the agent will bid .5 × 30 + .25 × 50 + .25 × 10 = $30 for the flash.

Notice that in this analysis of what to bid for the flash, the actual closing price of the flash is irrelevant. The proper bid depends only on the predicted price of the camera. To determine the proper bid for the camera, a similar analysis would be done using the predicted price distribution of the flash.

## 3. TAC

We instantiated our approach as an entry in the second Trading Agent Competition (TAC), as described in this section. Building on the success of TAC-00 held in July 2000 (Wellman, Wurman, O'Malley, Bangera, Lin, Reeves, & Walsh, 2001), TAC-01 included 19 agents from 9 countries (Wellman et al., 2003a). A key feature of TAC is that it required *autonomous bidding agents* to buy and sell *multiple interacting goods* in auctions of different types. It is designed as a benchmark problem in the complex and rapidly advancing domain of e-marketplaces, motivating researchers to apply unique approaches to a common task. By providing a clear-cut objective function, TAC also allows the competitors to focus their attention on the computational and game-theoretic aspects of the problem and leave aside the modeling and model validation issues that invariably loom large in real applications of automated agents to auctions (see Rothkopf & Harstad, 1994). Another feature of TAC is that it provides an academic forum for open comparison of agent bidding strategies in a complex scenario, as opposed to other complex scenarios, such as trading in real stock markets, in which practitioners are (understandably) reluctant to share their technologies.

A TAC game instance pits eight autonomous bidding agents against one another. Each TAC agent is a simulated travel agent with eight clients, each of whom would like to travel from TACtown to Tampa and back again during a 5-day period. Each client is characterized by a random set of preferences for the possible arrival and departure dates, hotel rooms, and entertainment tickets. To satisfy a client, an agent must construct a travel package for that client by purchasing airline tickets to and from TACtown and securing hotel reservations; it is possible to obtain additional bonuses by providing entertainment tickets as well. A TAC agent's score in a game instance is the difference between the sum of its clients' utilities for the packages they receive and the agent's total expenditure. We provide selected details about the game next; for full details on the design and mechanisms of the TAC server and TAC game, see `http://www.sics.se/tac`.

TAC agents buy flights, hotel rooms and entertainment tickets through auctions run from the TAC server at the University of Michigan. Each game instance lasts 12 minutes and includes a total of 28 auctions of 3 different types.

**Flights (8 auctions):** There is a separate auction for each type of airline ticket: to Tampa (*inflights*) on days 1–4 and from Tampa (*outflights*) on days 2–5. There is an unlimited supply of airline tickets, and every 24–32 seconds their ask price changes by from −$10

to \$$x$. $x$ increases linearly over the course of a game from 10 to $y$, where $y \in [10, 90]$ is chosen uniformly at random for each auction, and is unknown to the bidders. In all cases, tickets are priced between \$150 and \$800. When the server receives a bid at or above the ask price, the transaction is cleared immediately at the ask price and no resale is allowed.

**Hotel Rooms (8):** There are two different types of hotel rooms—the Tampa Towers (TT) and the Shoreline Shanties (SS)—each of which has 16 rooms available on days 1–4. The rooms are sold in a 16th-price *ascending* (English) auction, meaning that for each of the 8 types of hotel rooms, the 16 highest bidders get the rooms at the 16th highest price. For example, if there are 15 bids for TT on day 2 at \$300, 2 bids at \$150, and any number of lower bids, the rooms are sold for \$150 to the 15 high bidders plus one of the \$150 bidders (earliest received bid). The ask price is the current 16th-highest bid and transactions clear only when the auction closes. Thus, agents have no knowledge of, for example, the current highest bid. New bids must be higher than the current ask price. No bid withdrawal or resale is allowed, though the price of bids may be lowered provided the agent does not reduce the number of rooms it would win were the auction to close. One *randomly chosen* hotel auction closes at minutes 4–11 of the 12-minute game. Ask prices are changed only on the minute.

**Entertainment Tickets (12):** Alligator wrestling, amusement park, and museum tickets are each sold for days 1–4 in continuous double auctions. Here, agents can *buy and sell* tickets, with transactions clearing immediately when one agent places a buy bid at a price at least as high as another agent's sell price. Unlike the other auction types in which the goods are sold from a centralized stock, each agent starts with a (skewed) random endowment of entertainment tickets. The prices sent to agents are the *bid-ask spreads*, *i.e.*, the highest current bid price and the lowest current ask price (due to immediate clears, ask price is always greater than bid price). In this case, *bid withdrawal* and *ticket resale* are both permitted. Each agent gets blocks of 4 tickets of 2 types, 2 tickets of another 2 types, and no tickets of the other 8 types.

In addition to unpredictable market prices, other sources of variability from game instance to game instance are the client profiles assigned to the agents and the random initial allotment of entertainment tickets. Each TAC agent has eight clients with randomly assigned travel preferences. Clients have parameters for ideal arrival day, *IAD* (1–4); ideal departure day, *IDD* (2–5); hotel premium, *HP* (\$50–\$150); and entertainment values, *EV* (\$0–\$200) for each type of entertainment ticket.

The utility obtained by a client is determined by the travel package that it is given in combination with its preferences. To obtain a non-zero utility, the client must be assigned a *feasible* travel package consisting of an inflight on some arrival day *AD*, an outflight on a departure day *DD*, and hotel rooms of the *same type* (TT or SS) for the days in between (days $d$ such that $AD \leq d < DD$). At most one entertainment ticket of each type can be assigned, and no more than one on each day. Given a feasible package, the client's utility is defined as

$$1000 - travelPenalty + hotelBonus + funBonus$$

where





- $travelPenalty = 100(|AD - IAD| + |DD - IDD|)$

- $hotelBonus = HP$ if the client is in the TT, 0 otherwise.

- $funBonus =$ sum of $EVs$ for assigned entertainment tickets.

A TAC agent's *score* is the sum of its clients' utilities in the optimal allocation of its goods (computed by the TAC server) minus its expenditures. The client preferences, allocations, and resulting utilities from a sample game are shown in Tables 3 and 4.

| Client | $IAD$ | $IDD$ | $HP$ | $AW$ | $AP$ | $MU$ |
|--------|-------|-------|------|------|------|------|
| 1 | Day 2 | Day 5 | 73 | 175 | 34 | 24 |
| 2 | Day 1 | Day 3 | 125 | 113 | 124 | 57 |
| 3 | Day 4 | Day 5 | 73 | 157 | 12 | 177 |
| 4 | Day 1 | Day 2 | 102 | 50 | 67 | 49 |
| 5 | Day 1 | Day 3 | 75 | 12 | 135 | 110 |
| 6 | Day 2 | Day 4 | 86 | 197 | 8 | 59 |
| 7 | Day 1 | Day 5 | 90 | 56 | 197 | 162 |
| 8 | Day 1 | Day 3 | 50 | 79 | 92 | 136 |

Table 3: ATTac-2001's client preferences from an actual game. $AW$, $AP$, and $MU$ are $EVs$ for alligator wrestling, amusement park, and museum respectively.

| Client | $AD$ | $DD$ | Hotel | Ent'ment | Utility |
|--------|------|------|-------|----------|---------|
| 1 | Day 2 | Day 5 | SS | AW4 | 1175 |
| 2 | Day 1 | Day 2 | TT | AW1 | 1138 |
| 3 | Day 3 | Day 5 | SS | MU3, AW4 | 1234 |
| 4 | Day 1 | Day 2 | TT | None | 1102 |
| 5 | Day 1 | Day 2 | TT | AP1 | 1110 |
| 6 | Day 2 | Day 3 | TT | AW2 | 1183 |
| 7 | Day 1 | Day 5 | SS | AF2, AW3, MU4 | 1415 |
| 8 | Day 1 | Day 2 | TT | MU1 | 1086 |

Table 4: ATTac-2001's client allocations and utilities from the same actual game as that in Table 3. Client 1's "4" under "Ent'ment" indicates on day 4.

The rules of TAC-01 are largely identical to those of TAC-00, with three important exceptions:

1. In TAC-00, flight prices did not tend to increase;

2. In TAC-00, hotel auctions usually all closed at the end of the game;

3. In TAC-00, entertainment tickets were distributed uniformly to all agents

While relatively minor on the surface, these changes significantly enriched the strategic complexity of the game. Stone and Greenwald (2003) detail agent strategies from TAC-00.





TAC-01 was organized as a series of four competition phases, culminating with the semifinals and finals on October 14, 2001 at the EC-01 conference in Tampa, Florida. First, the qualifying round, consisting of about 270 games per agent, served to select the 16 agents that would participate in the semifinals. Second, the seeding round, consisting of about 315 games per agent, was used to divide these agents into two groups of eight. After the semifinals, on the morning of the 14th consisting of 11 games in each group, four teams from each group were selected to compete in the finals during that same afternoon. The finals are summarized in Section 6.

TAC is not designed to be fully realistic in the sense that an agent from TAC is not immediately deployable in the real world. For one thing, it is unrealistic to assume that an agent would have complete, reliable access to all clients' utility functions (or even that the client would!); typically, some sort of preference elicitation procedure would be required (e.g. Boutilier, 2002). For another, the auction mechanisms are somewhat contrived for the purposes of creating an interesting, yet relatively simple game. However, each mechanism *is* representative of a class of auctions that is used in the real world. And it is not difficult to imagine a future in which agents do need to bid in decentralized, related, yet varying auctions for similarly complex packages of goods.

## 4. Hotel Price Prediction

As discussed earlier, a central part of our strategy depends on the ability to predict prices, particularly hotel prices, at various points in the game. To do this as accurately as possible, we used machine-learning techniques that would examine the hotel prices actually paid in previous games to predict prices in future games. This section discusses this part of our strategy in detail, including a new boosting-based algorithm for conditional density estimation.

There is bound to be considerable uncertainty regarding hotel prices since these depend on many unknown factors, such as the time at which the hotel room will close, who the other agents are, what kind of clients have been assigned to each agent, etc. Thus, *exactly* predicting the price of a hotel room is hopeless. Instead, we regard the closing price as a random variable that we need to estimate, conditional on our current state of knowledge (*i.e.*, number of minutes remaining in the game, ask price of each hotel, flight prices, etc.). We might then attempt to predict this variable's conditional expected value. However, our strategy requires that we not only predict expected value, but that we also be able to estimate the *entire* conditional distribution so that we can *sample* hotel prices.

To set this up as a learning problem, we gathered a set of training examples from previously played games. We defined a set of features for describing each example that together are meant to comprise a snap-shot of all the relevant information available at the time each prediction is made. All of the features we used are real valued; a couple of the features can have a special value $\perp$ indicating "value unknown." We used the following basic features:

- The number of minutes remaining in the game.

- The price of each hotel room, *i.e.*, the current ask price for rooms that have not closed or the actual selling price for rooms that have closed.





- The closing time of each hotel room. Note that this feature is defined even for rooms that have not yet closed, as explained below.

- The prices of each of the flights.

To this basic list, we added a number of redundant variations, which we thought might help the learning algorithm:

- The closing price of hotel rooms that have closed (or ⊥ if the room has not yet closed).

- The current ask price of hotel rooms that have not closed (or ⊥ if the room has already closed).

- The closing time of each hotel room minus the closing time of the room whose price we are trying to predict.

- The number of minutes from the current time until each hotel room closes.

During the seeding rounds, it was impossible to know during play who our opponents were, although this information was available at the end of each game, and therefore during training. During the semifinals and finals, we did know the identities of all our competitors. Therefore, in preparation for the semifinals and finals, we added the following features:

- The number of players playing (ordinarily eight, but sometimes fewer, for instance if one or more players crashed).

- A bit for each player indicating whether or not that player participated in this game.

We trained specialized predictors for predicting the price of each type of hotel room. In other words, one predictor was specialized for predicting only the price of TT on day 1, another for predicting SS on day 2, etc. This would seem to require eight separate predictors. However, the tournament game is naturally symmetric about its middle in the sense that we can create an equivalent game by exchanging the hotel rooms on days 1 and 2 with those on days 4 and 3 (respectively), and by exchanging the inbound flights on days 1, 2, 3 and 4 with the outbound flights on days 5, 4, 3 and 2 (respectively). Thus, with appropriate transformations, the outer days (1 and 4) can be treated equivalently, and likewise for the inner days (2 and 3), reducing the number of specialized predictors by half.

We also created specialized predictors for predicting in the first minute after flight prices had been quoted but prior to receiving any hotel price information. Thus, a total of eight specialized predictors were built (for each combination of TT versus SS, inner versus outer day, and first minute versus not first minute).

We trained our predictors to predict not the actual closing price of each room per se, but rather how much the price would increase, i.e., the difference between the closing price and the current price. We thought that this might be an easier quantity to predict, and, because our predictor never outputs a negative number when trained on nonnegative data, this approach also ensures that we never predict a closing price below the current bid.

From each of the previously played games, we were able to extract many examples. Specifically, for each minute of the game and for each room that had not yet closed, we





extracted the values of all of the features described above at that moment in the game, plus the actual closing price of the room (which we are trying to predict).

Note that during training, there is no problem extracting the closing times of all of the rooms. During the actual play of a game, we do not know the closing times of rooms that have not yet closed. However, we do know the exact probability distribution for closing times of all of the rooms that have not yet closed. Therefore, to sample a vector of hotel prices, we can first sample according to this distribution over closing times, and then use our predictor to sample hotel prices using these sampled closing times.

## 4.1 The Learning Algorithm

Having described how we set up the learning problem, we are now ready to describe the learning algorithm that we used. Briefly, we solved this learning problem by first reducing to a multiclass, multi-label classification problem (or alternatively a multiple logistic regression problem), and then applying boosting techniques developed by Schapire and Singer (1999, 2000) combined with a modification of boosting algorithms for logistic regression proposed by Collins, Schapire and Singer (2002). The result is a new machine-learning algorithm for solving conditional density estimation problems, described in detail in the remainder of this section. Table 5 shows pseudo-code for the entire algorithm.

Abstractly, we are given pairs $(x_1, y_1), \ldots, (x_m, y_m)$ where each $x_i$ belongs to a space $X$ and each $y_i$ is in $\mathbb{R}$. In our case, the $x_i$'s are the auction-specific feature vectors described above; for some $n$, $X \subseteq (\mathbb{R} \cup \{\perp\})^n$. Each target quantity $y_i$ is the difference between closing price and current price. Given a new $x$, our goal is to estimate the conditional distribution of $y$ given $x$.

We proceed with the working assumption that all training and test examples $(x, y)$ are i.i.d. (i.e, drawn independently from identical distributions). Although this assumption is false in our case (since the agents, including ours, are changing over time), it seems like a reasonable approximation that greatly reduces the difficulty of the learning task.

Our first step is to reduce the estimation problem to a classification problem by breaking the range of the $y_i$'s into bins:

$$[b_0, b_1), [b_1, b_2), \ldots, [b_k, b_{k+1}]$$

for some breakpoints $b_0 < b_1 < \cdots < b_k \leq b_{k+1}$ where for our problem, we chose $k = 50$.[7] The endpoints $b_0$ and $b_{k+1}$ are chosen to be the smallest and largest $y_i$ values observed during training. We choose the remaining breakpoints $b_1, \ldots, b_k$ so that roughly an equal number of training labels $y_i$ fall into each bin. (More technically, breakpoints are chosen so that the entropy of the distribution of bin frequencies is maximized.)

For each of the breakpoints $b_j$ $(j = 1, \ldots, k)$, our learning algorithm attempts to estimate the probability that a new $y$ (given $x$) will be at least $b_j$. Given such estimates $p_j$ for each $b_j$, we can then estimate the probability that $y$ is in the bin $[b_j, b_{j+1})$ by $p_{j+1} - p_j$ (and we can then use a constant density within each bin). We thus have reduced the problem to one of estimating multiple conditional Bernoulli variables corresponding to the event

---

7. We did not experiment with varying $k$, but expect that the algorithm is not sensitive to it for sufficiently large values of $k$.





---

**Input:** $(x_1, y_1), \ldots, (x_m, y_m)$ where $x_i \in X$, $y_i \in \mathbb{R}$
  positive integers $k$ and $T$

**Compute breakpoints:** $b_0 < b_1 < \cdots < b_{k+1}$ where
- $b_0 = \min_i y_i$
- $b_{k+1} = \max_i y_i$
- $b_1, \ldots, b_k$ chosen to minimize $\sum_{j=0}^{k} q_j \ln q_j$ where $q_0, \ldots, q_k$ are fraction of $y_i$'s in $[b_0, b_1), [b_1, b_2), \ldots, [b_k, b_{k+1}]$ (using dynamic programing)

**Boosting:**
- for $t = 1, \ldots T$:
  - compute weights $W_t(i, j) = \dfrac{1}{1 + e^{s_j(y_i) f_t(x_i, j)}}$

    where $s_j(y)$ is as in Eq. (10)
  - use $W_t$ to obtain base function $h_t : X \times \{1, \ldots, k\} \to \mathbb{R}$ minimizing

    $\displaystyle\sum_{i=1}^{m} \sum_{j=1}^{k} W_t(i, j) e^{-s_j(y_i) h_t(x_i, j)}$ over all decision rules $h_t$ considered. The decision rules can take any form. In our work, we use "decision stumps," or simple thresholds on one of the features.

**Output sampling rule:**
- let $f = \displaystyle\sum_{t=1}^{T} h_t$
- let $f' = (\overline{f} + \underline{f})/2$ where

$$
\begin{aligned}
\overline{f}(x, j) &= \max\{f(x, j') : j \le j' \le k\} \\
\underline{f}(x, j) &= \min\{f(x, j') : 1 \le j' \le j\}
\end{aligned}
$$

- to sample, given $x \in X$
  - let $p_j = \dfrac{1}{1 + e^{-f'(x, j)}}$
  - let $p_0 = 1, p_{k+1} = 0$
  - choose $j \in \{0, \ldots, k\}$ randomly with probability $p_j - p_{j+1}$
  - choose $y$ uniformly at random from $[b_j, b_{j+1}]$
  - output $y$

---

Table 5: The boosting-based algorithm for conditional density estimation.

$y \ge b_j$, and for this, we use a logistic regression algorithm based on boosting techniques as described by Collins et al. (2002).

Our learning algorithm constructs a real-valued function $f : X \times \{1, \ldots, k\} \to \mathbb{R}$ with the interpretation that

$$
\frac{1}{1 + \exp(-f(x, j))} \tag{9}
$$

is our estimate of the probability that $y \ge b_j$, given $x$. The negative log likelihood of the conditional Bernoulli variable corresponding to $y_i$ being above or below $b_j$ is then

$$
\ln\left(1 + e^{-s_j(y_i) f(x_i, j)}\right)
$$





where

$$s_j(y) = \begin{cases} +1 & \text{if } y \geq b_j \\ -1 & \text{if } y < b_j. \end{cases} \tag{10}$$

We attempt to minimize this quantity for all training examples $(x_i, y_i)$ and all breakpoints $b_j$. Specifically, we try to find a function $f$ minimizing

$$\sum_{i=1}^{m} \sum_{j=1}^{k} \ln \left( 1 + e^{-s_j(y_i) f(x_i, j)} \right).$$

We use a boosting-like algorithm described by Collins et al. (2002) for minimizing objective functions of exactly this form. Specifically, we build the function $f$ in rounds. On each round $t$, we add a new base function $h_t : X \times \{1, \ldots, k\} \to \mathbb{R}$. Let

$$f_t = \sum_{t'=1}^{t-1} h_{t'}$$

be the accumulating sum. Following Collins, Schapire and Singer, to construct each $h_t$, we first let

$$W_t(i, j) = \frac{1}{1 + e^{s_j(y_i) f_t(x_i, j)}}$$

be a set of weights on example-breakpoint pairs. We then choose $h_t$ to minimize

$$\sum_{i=1}^{m} \sum_{j=1}^{k} W_t(i, j) e^{-s_j(y_i) h_t(x_i, j)} \tag{11}$$

over some space of "simple" base functions $h_t$. For this work, we considered all "decision stumps" $h$ of the form

$$h(x, j) = \begin{cases} A_j & \text{if } \phi(x) \geq \theta \\ B_j & \text{if } \phi(x) < \theta \\ C_j & \text{if } \phi(x) = \perp \end{cases}$$

where $\phi(\cdot)$ is one of the features described above, and $\theta$, $A_j$, $B_j$ and $C_j$ are all real numbers. In other words, such an $h$ simply compares one feature $\phi$ to a threshold $\theta$ and returns a vector of numbers $h(x, \cdot)$ that depends only on whether $\phi(x)$ is unknown ($\perp$), or above or below $\theta$. Schapire and Singer (2000) show how to efficiently search for the best such $h$ over all possible choices of $\phi$, $\theta$, $A_j$, $B_j$ and $C_j$. (We also employed their technique for "smoothing" $A_j$, $B_j$ and $C_j$.)

When computed by this sort of iterative procedure, Collins et al. (2002) prove the asymptotic convergence of $f_t$ to the minimum of the objective function in Equation (11) over all linear combinations of the base functions. For this problem, we fixed the number of rounds to $T = 300$. Let $f = f_{T+1}$ be the final predictor.

As noted above, given a new feature vector $x$, we compute $p_j$ as in Equation (9) to be our estimate for the probability that $y \geq b_j$, and we let $p_0 = 1$ and $p_{k+1} = 0$. For this to make sense, we need $p_1 \geq p_2 \geq \cdots \geq p_k$, or equivalently, $f(x, 1) \geq f(x, 2) \geq \cdots \geq f(x, k)$, a condition that may not hold for the learned function $f$. To force this condition, we replace $f$ by a reasonable (albeit heuristic) approximation $f'$ that is nonincreasing in $j$, namely,





$f' = (\overline{f} + \underline{f})/2$ where $\overline{f}$ (respectively, $\underline{f}$) is the pointwise minimum (respectively, maximum) of all nonincreasing functions $g$ that everywhere upper bound $f$ (respectively, lower bound $f$).

With this modified function $f'$, we can compute modified probabilities $p_j$. To sample a single point according to the estimated distribution on $\mathbb{R}$ associated with $f'$, we choose bin $[b_j, b_{j+1})$ with probability $p_j - p_{j+1}$, and then select a point from this bin uniformly at random. Expected value according to this distribution is easily computed as

$$\sum_{j=0}^{k} (p_j - p_{j+1}) \left( \frac{b_{j+1} + b_j}{2} \right).$$

Although we present results using this algorithm in the trading agent context, we did not test its performance on more general learning problems, nor did we compare to other methods for conditional density estimation, such as those studied by Stone (1994). This clearly should be an area for future research.

## 5. ATTac-2001

Having described hotel price prediction in detail, we now present the remaining details of ATTac-2001's algorithm. We begin with a brief description of the goods allocator, which is used as a subroutine throughout the algorithm. We then present the algorithm in a top-down fashion.

### 5.1 Starting Point

A core subproblem for TAC agents is the allocation problem: finding the most profitable allocation of goods to clients, $G^*$, given a set of owned goods and prices for all other goods. The allocation problem corresponds to finding $\mathrm{opt}(H, i, Y)$ in Equation 3. We denote the value of $G^*$ (i.e., the score one would attain with $G^*$) as $v(G^*)$. The general allocation problem is NP-complete, as it is equivalent to the set-packing problem (Garey & Johnson, 1979). However it can be solved tractably in TAC via integer linear programming (Stone et al., 2001).

The solution to the integer linear program is a value-maximizing allocation of owned resources to clients along with a list of resources that need to be purchased. Using the linear programming package "LPsolve", ATTac-2001 is usually able to find the globally optimal solution in under 0.01 seconds on a 650 MHz Pentium II. However, since integer linear programming is an NP-complete problem, some inputs can lead to a great deal of search over the integrality constraints, and therefore significantly longer solution times. When only $v(G^*)$ is needed (as opposed to $G^*$ itself), the upper bound produced by LPsolve prior to the search over the integrality constraints, known as the LP relaxation, can be used as an estimate. The LP relaxation can always be generated very quickly.

Note that this is not by any means the only possible formulation of the allocation problem. Greenwald and Boyan (2001) studied a fast, heuristic search variant and found that it performed extremely well on a collection of large, random allocation problems. Stone et al. (2001) used a randomized greedy strategy as a fallback for the cases in which the linear program took too long to solve.

## 5.2 Overview

Table 6 shows a high-level overview of ATTac-2001. The italicized portions are described in the remainder of this section.

---

**When the first flight quotes are posted:**
- Compute $G^*$ with current holdings and *expected prices*
- Buy the flights in $G^*$ for which *expected cost of postponing commitment* exceeds the *expected benefit of postponing commitment*

**Starting 1 minute before each hotel close:**
- Compute $G^*$ with current holdings and *expected prices*
- Buy the flights in $G^*$ for which *expected cost of postponing commitment* exceeds *expected benefit of postponing commitment* (30 seconds)
- Bid *hotel room expected marginal values* given holdings, new flights, and *expected hotel purchases* (30 seconds)

**Last minute:** Buy remaining flights as needed by $G^*$

**In parallel (continuously):** Buy/sell entertainment tickets based on their *expected values*

---

Table 6: ATTac-2001's high-level algorithm. The italicized portions are described in the remainder of this section.

## 5.3 Cost of Additional Rooms

Our hotel price predictor described in Section 4 assumes that ATTac-2001's bids do not affect the ultimate closing price (Assumption 3 from Section 2). This assumption holds in a large economy. However in TAC, each hotel auction involved 8 agents competing for 16 hotel rooms. Therefore, the actions of each agent had an appreciable effect on the clearing price: the more hotel rooms an agent attempted to purchase, the higher the clearing price would be, all other things being equal. This effect needed to be taken into account when solving the basic allocation problem.

The simplified model used by ATTac-2001 assumed that the $n$th highest bid in a hotel auction was roughly proportional to $c^{-n}$ (over the appropriate range of $n$) for some $c \geq 1$. Thus, if the predictor gave a price of $p$, ATTac-2001 only used this for purchasing two hotel rooms (the "fair" share of a single agent of the 16 rooms), and adjusted prices for other quantities of rooms by using $c$.

For example, ATTac-2001 would consider the cost of obtaining 4 rooms to be $4pc^2$. One or two rooms each cost $p$, but 3 each cost $pc$, 4 each cost $pc^2$, 5 each cost $pc^3$, etc. So in total, 2 rooms cost $2p$, while 4 cost $4pc^2$. The reasoning behind this procedure is that if ATTac-2001 buys two rooms — its fair share given that there are 16 rooms and 8 agents, then the 16th highest bid (ATTac-2001's 2 bids in addition to 14 others) sets the price. But if ATTac-2001 bids on an additional unit, the previous 15th highest bid becomes the price-setting bid: the price for all rooms sold goes up from $p$ to $pc$.

The constant $c$ was calculated from the data of several hundred games during the seeding round. In each hotel auction, the ratio of the 14th and 18th highest bids (reflecting the most relevant range of $n$) was taken as an estimate of $c^4$, and the (geometric) mean of the resulting estimates was taken to obtain $c = 1.35$.

The LP allocator takes these price estimates into account when computing $G^*$ by assigning higher costs to larger purchase volumes, thus tending to spread out ATTac-2001's demand over the different hotel auctions.

In ATTac-2001, a few heuristics were applied to the above procedure to improve stability and to avoid pathological behavior: prices below \$1 were replaced by \$1 in estimating $c$; $c = 1$ was used for purchasing fewer than two hotel rooms; hotel rooms were divided into early closing and late closing (and cheap and expensive) ones, and the $c$ values from the corresponding subsets of auctions of the seeding rounds were used in each case.

## 5.4 Hotel Expected Marginal Values

Using the hotel price prediction module as described in Section 4, coupled with a model of its own effect on the economy, ATTac-2001 is equipped to determine its bids for hotel rooms.

Every minute, for each hotel auction that is still open, ATTac-2001 assumes that auction will close next and computes the marginal value of that hotel room given the predicted closing prices of the other rooms. If the auction does not close next, then it assumes that it will have a chance to revise its bids. Since these predicted prices are represented as distributions of possible future prices, ATTac-2001 samples from these distributions and averages the marginal values to obtain an expected marginal value. Using the full minute between closing times for computation (or 30 seconds if there are still flights to consider too), ATTac-2001 divides the available time among the different open hotel auctions and generates as many price samples as possible for each hotel room. In the end, ATTac-2001 bids the expected marginal values for each of the rooms.

The algorithm is described precisely and with explanation in Table 7.

One additional complication regarding hotel auctions is that, contrary to one of our assumptions in Section 2.2 (Assumption 4), bids are not fully retractable: they can only be changed to \$1 above the current ask price. In the case that there are current active bids for goods that ATTac-2001 no longer wants that are less than \$1 above the current ask price, it may be advantageous to refrain from changing the bid in the hopes that the ask price will surpass them: that is, the current bid may have a higher expected value than the best possible new bid. To address this issue, ATTac-2001 samples from the learned price distribution to find the average expected values of the current and potential bids, and only enters a new bid in the case that the potential bid is better.

## 5.5 Expected Cost/Benefit of Postponing Commitment

ATTac-2001 makes flight bidding decisions based on a cost-benefit analysis: in particular, ATTac-2001 computes the incremental cost of postponing bidding for a particular flight versus the value of delaying commitment. In this section, we describe the determination of the cost of postponing the bid.

Due to difficulties that compounded with more sophisticated approaches, ATTac-2001 used the following very simple model for estimating the price of a flight ticket at a given future time. It is evident from the formulation that—given $y$—the expected price increase from time 0 to time $t$ was very nearly of the form $Mt^2$ for some $M$. It was also clear that

- For each hotel (in order of increasing expected price):
- Repeat until time bound
  1. Generate a random hotel closing order (only over open hotels)
  2. *Sample* closing prices from predicted hotel price distributions
  3. Given these closing prices, compute $V_0, V_1, \ldots V_n$
     - $V_i \equiv v(G^*)$ if owning $i$ of the hotel
     - Estimate $v(G^*)$ with LP relaxation
     - Assume that no additional hotel rooms of this type can be bought
     - For other hotels, assume outstanding bids above sampled price are already owned (*i.e.*, they cannot be withdrawn).
     - Note that $V_0 \leq V_1 \leq \ldots \leq V_n$: the values are monotonically increasing since having more goods cannot be worse in terms of possible allocations
- The value of the $i$th copy of the room is the mean of $V_i - V_{i-1}$ over all the samples.
- Note further that $V_1 - V_0 \geq V_2 - V_1 \ldots \geq V_n - V_{n-1}$: the value differences are monotonically decreasing since each additional room will be assigned to the client who can derive the most value from it.
- Bid for one room at the value of the $i$th copy of the room for all $i$ such that the value is at least as much as the current price. Due to the monotonicity noted in the step above, no matter what the closing price, the desired number of rooms at that price will be purchased.

Table 7: The algorithm for generating bids for hotel rooms.

as long as the price did not hit the artificial boundaries at \$150 and \$800, the constant $M$ must depend linearly on $y - 10$. This linear dependence coefficient was then estimated from several hundred flight price evolutions during the qualifying round. Thus, for this constant $m$, the expected price increase from time $t$ to time $T$ was $m(T^2 - t^2)(y - 10)$. When a price prediction was needed, this formula was first used for the first and most recent actual price observations to obtain a guess for $y$, and then this $y$ was used in the formula again to estimate the future price. No change was predicted if the formula yielded a price decrease.

This approach suffers from systemic biases of various kinds (mainly due to the fact that the variance of price changes gets relatively smaller over longer periods of time), but was thought to be accurate enough for its use, which was to predict whether or not the ticket can be expected to get significantly more expensive over the next few minutes.

In practice, during TAC-01, ATTac-2001 started with the *flight-lookahead* parameter set to 3 (*i.e.*, cost of postponing is the average of the predicted flight costs 1, 2, and 3 minutes in the future). However, this parameter was changed to 2 by the end of the finals in order to cause ATTac-2001 to delay its flight commitments further.

### 5.5.1 EXPECTED BENEFIT OF POSTPONING COMMITMENT

Fundamentally, the benefit of postponing commitments to flights is that additional information about the eventual hotel prices becomes known. Thus, the benefit of postponing commitment is computed by sampling possible future price vectors and determining, on average, how much better the agent could do if it bought a different flight instead of the one in question. If it is optimal to buy the flight in all future scenarios, then there is no value in delaying the commitment and the flight is purchased immediately. However, if there are many scenarios in which the flight is not the best one to get, the purchase is more likely to be delayed.

The algorithm for determining the benefit of postponing commitment is similar to that for determining the marginal value of hotel rooms. It is detailed, with explanation, in Table 8.

---

- Assume we're considering buying $n$ flights of a given type

- Repeat until time bound
    1. Generate a random hotel closing order (open hotels)
    2. *Sample* closing prices from predicted price distributions (open hotels)
    3. Given these closing prices compute $V_0, V_1, \ldots V_n$
        - $V_i = v(G^*)$ if forced to buy $i$ of the flight
        - Estimate $v(G^*)$ with LP relaxation
        - Assume more flights can be bought at the *current price*
        - Note that $V_0 \geq V_1 \geq \ldots \geq V_n$ since it is never worse to retain extra flexibility.

- The value of waiting to buy copy $i$ is the mean of $V_i - V_{i-1}$ over all the samples. If all price samples lead to the conclusion that the $i$th flight should be bought, then $V_i = V_{i-1}$ and there is no benefit to postponing commitment.

---

Table 8: The algorithm for generating value of postponing flight commitments.

## 5.6 Entertainment Expected Values

The core of ATTac-2001's entertainment-ticket-bidding strategy is again a calculation of the expected marginal values of each ticket. For each ticket, ATTac-2001 computes the expected value of having one more and one fewer of the ticket. These calculations give bounds on the bid and ask prices it is willing to post. The actual bid and ask prices are a linear function of time remaining in the game: ATTac-2001 settles for a smaller and smaller profit from ticket transactions as the game goes on. Details of the functions of bid and ask price as a function of game time and ticket value remained unchanged from ATTac-2000 (Stone et al., 2001).

Details of the entertainment-ticket expected marginal utility calculations are given in Table 9.

## 6. Results

This section presents empirical results demonstrating the effectiveness of the ATTac-2001 strategy. First, we summarize its performance in the 2001 and 2002 Trading Agent Competitions (TACs). These summaries provide evidence of the strategy's overall effectiveness, but, due to the small number of games in the competitions, are anecdotal rather than scientifically conclusive. We then present controlled experiments that provide more conclusive evidence of the utility of our decision theoretic and learning approaches embedded within ATTac-2001.

- Assume $n$ of a given ticket type are currently owned

- Repeat until time bound
  1. Generate a random hotel closing order (open hotels)
  2. *Sample* closing prices from predicted price distributions (open hotels)
  3. Given these closing prices compute $V_{n-1}, V_n, V_{n+1}$
     - $V_i = v(G^*)$ if own $i$ of the ticket
     - Estimate $v(G^*)$ with LP relaxation
     - Assume no other tickets can be bought or sold
     - Note that $V_{n-1} \leq V_n \leq V_{n+1}$ since it is never worse to own extra tickets.

- The value of buying a ticket is the mean of $V_{n+1} - V_n$ over all the samples; the value of selling is the mean of $V_n - V_{n-1}$.

- Since tickets are considered sequentially, if the determined buy or sell bid leads to a price that would clear according to the current quotes, assume the transaction goes through before computing the values of buying and selling other ticket types.

Table 9: The algorithm for generating value of entertainment tickets.

## 6.1 TAC-01 Competition

Of the 19 teams that entered the qualifying round, ATTac-2001 was one of eight agents to make it to the finals on the afternoon of October 14th, 2001. The finals consisted of 24 games among the same eight agents. Right from the beginning, it became clear that livingagents (Fritschi & Dorer, 2002) was the team to beat in the finals. They jumped to an early lead in the first two games, and by eight games into the round, they were more than 135 points per game ahead of their closest competitor (SouthamptonTAC, He & Jennings, 2002). 16 games into the round, they were more than 250 points ahead of their two closest competitors (ATTac-2001 and whitebear).

From that point, ATTac-2001, which was continually retraining its price predictors based on recent games, began making a comeback. By the time the last game was to be played, it was only an average of 22 points per game behind livingagents. It thus needed to beat livingagents by 514 points in the final game to overtake it, well within the margins observed in individual game instances. As the game completed, ATTac-2001's score of 3979 was one of the first scores to be posted by the server. The other agents' scores were reported one by one, until only the livingagents score was left. After agonizing seconds (at least for us), the TAC server posted a final game score of 4626, resulting in a win for livingagents.

After the competition, the TAC team at the University of Michigan conducted a regression analysis of the effects of the client profiles on agent scores. Using data from the seeding rounds, it was determined that agents did better when their clients had:

1. fewer total preferred travel days;

2. higher total entertainment values; and

3. a higher ratio of outer days (1 and 4) to inner (2 and 3) in preferred trip intervals.

Based on these significant measures, the games in the finals could be handicapped based on each agents' aggregate client profiles. Doing so indicated that livingagents' clients were much easier to satisfy than those of ATTac-2001, giving ATTac-2001 the highest handicapped score. The final scores, as well as the handicapped scores, are shown in Table 10. Complete results and affiliations are available from http://tac.eecs.umich.edu.

| Agent | Mean | Handicapped score |
|---|---|---|
| ATTac-2001 | 3622 | 4154 |
| livingagents | 3670 | 4094 |
| whitebear | 3513 | 3931 |
| Urlaub01 | 3421 | 3909 |
| Retsina | 3352 | 3812 |
| CaiserSose | 3074 | 3766 |
| SouthamptonTAC | 3253 | 3679 |
| TacsMan | 2859 | 3338 |

Table 10: Scores during the finals. Each agent played 24 games. Southampton's score was adversely affected by a game in which their agent crashed after buying many flights but no hotels, leading to a loss of over 3000 points. Discarding that game results in an average score of 3531.

## 6.2 TAC-02 Competition

A year after the TAC-01 competition, ATTac-2001 was re-entered in the TAC-02 competition using the models trained at the end of TAC-01. Specifically, the price predictors were left unchanged throughout (no learning). The seeding round included 19 agents, each playing 440 games over the course of about 2 weeks. ATTac-2001 was the top-scoring agent in this round, as shown in Table 11. Scores in the seeding round were weighted so as to emphasize later results over earlier results: scores on day $n$ of the seeding round were given a weight of $n$. This practice was designed to encourage experimentation early in the round. The official ranking in the competitions was based on the mean score after ignoring each agent's worst 10 results so as to allow for occasional program crashes and network problems.

On the one hand, it is striking that ATTac-2001 was able to finish so strongly in a field of agents that had presumably improved over the course of the year. On the other hand, most agents were being tuned, for better and for worse, while ATTac-2001 was consistent throughout. In particular, we are told that SouthamptonTAC experimented with its approach during the later days of the round, perhaps causing it to fall out of the lead (by weighted score) in the end. During the 14-game semifinal heat, ATTac-2001, which was now restored with its learning capability and retrained over the data from the 2002 seeding round, finished 6th out of 8 thereby failing to reach the finals.

There are a number of possible reasons for this sudden failure. One relatively mundane explanation is that the agent had to change computational environments between the seeding rounds and the finals, and there may have been a bug or computational resource constraint introduced. Another possibility is that due to the small number of games in the semifinals, ATTac-2001 simply got unlucky with respect to clients and the interaction of opponent strategies. However, it is also plausible that the training data from the 2002 qualifying and seeding round data was less representative of the 2002 finals than the was the training data from 2001; and/or that the competing agents improved significantly over the seeding round while ATTac-2001 remained unchanged. The TAC team at the University of Michigan has done a study of the price predictors of several 2002 TAC agents that suggests that the bug hypothesis is most plausible: the ATTac-2001 predictor from 2001 outperforms all other predictors from 2002 on the data from the 2002 semifinals and finals; and one other agent that uses the 2002 data did produce good predictions based on that data (Wellman, Reeves, Lochner, & Vorobeychik, 2003b).[8]

| Agent | Mean | Weighted, dropped worst 10 |
|---|---|---|
| ATTac-2001 | 3050 | 3131 |
| SouthamptonTAC | 3100 | 3129 |
| UMBCTAC | 2980 | 3118 |
| livingagents | 3018 | 3091 |
| cuhk | 2998 | 3055 |
| Thalis | 2952 | 3000 |
| whitebear | 2945 | 2966 |
| RoxyBot | 2738 | 2855 |

Table 11: Top 8 scores during the seeding round of TAC-02. Each agent played 440 games, with its worst 10 games ignored when computing the rankings.

## 6.3 Controlled Experiments

ATTac-2001's success in the TAC-01 competition demonstrates its effectiveness as a complete system. However, since the competing agents differed along several dimensions, the competition results cannot isolate the successful approaches. In this section, we report controlled experiments designed to test the efficacy of ATTac-2001's machine-learning approach to price prediction.

### 6.3.1 VARYING THE PREDICTOR

In the first set of experiments, we attempted to determine how the quality of ATTac-2001's hotel price predictions affects its performance. To this end, we devised seven price prediction schemes, varying considerably in sophistication and inspired by approaches taken by other TAC competitors, and incorporated these schemes into our agent. We then played these seven agents against one another repeatedly, with regular retraining as described below.

Following are the seven hotel prediction schemes that we used, in decreasing order of sophistication:

---

8. Indeed, in the TAC-03 competition, ATTac-2001 was entered using the trained models from 2001, and it won the competition, suggesting further that the failure in 2002 was due to a problem with the learned models that were used during the finals in 2002.

- ATTac-2001$_s$: This is the "full-strength" agent based on boosting that was used during the tournament. (The $s$ denotes sampling.)

- Cond'lMean$_s$: This agent samples prices from the empirical distribution of prices from previously played games, conditioned only on the closing time of the hotel room (a subset of of the features used by ATTac-2001$_s$). In other words, it collects all historical hotel prices and breaks them down by the time at which the hotel closed (as well as room type, as usual). The price predictor then simply samples from the collection of prices corresponding to the given closing time.

- SimpleMean$_s$: This agent samples prices from the empirical distribution of prices from previously played games, without regard to the closing time of the hotel room (but still broken down by room type). It uses a subset of the features used by Cond'lMean$_s$.

- ATTac-2001$_{ev}$, Cond'lMean$_{ev}$, SimpleMean$_{ev}$: These agents predict in the same way as their corresponding predictors above, but instead of returning a random sample from the estimated distribution of hotel prices, they deterministically return the expected value of the distribution. (The $ev$ denotes expected value, as introduced in Section 2.1.)

- CurrentBid: This agent uses a very simple predictor that always predicts that the hotel room will close at its current price.

In every case, whenever the price predictor returns a price that is below the current price, we replace it with the current price (since prices cannot go down).

In our experiments, we added as an eighth agent EarlyBidder, inspired by the livingagents agent. EarlyBidder used SimpleMean$_{ev}$ to predict closing prices, determined an optimal set of purchases, and then placed bids for these goods at sufficiently high prices to ensure that they would be purchased ($1001 for all hotel rooms, just as livingagents did in TAC-01) right after the first flight quotes. It then never revised these bids.

Each of these agents require training, i.e., data from previously played games. However, we are faced with a sort of "chicken and egg" problem: to run the agents, we need to first train the agents using data from games in which they were involved, but to get this kind of data, we need to first run the agents. To get around this problem, we ran the agents in phases. In Phase I, which consisted of 126 games, we used training data from the seeding, semifinals and finals rounds of TAC-01. In Phase II, lasting 157 games, we retrained the agents once every six hours using all of the data from the seeding, semifinals and finals rounds as well as all of the games played in Phase II. Finally, in Phase III, lasting 622 games, we continued to retrain the agents once every six hours, but now using only data from games played during Phases I and II, and not including data from the seeding, semifinals and finals rounds.

Table 12 shows how the agents performed in each of these phases. Much of what we observe in this table is consistent with our expectations. The more sophisticated boosting-based agents (ATTac-2001$_s$ and ATTac-2001$_{ev}$) clearly dominated the agents based on simpler prediction schemes. Moreover, with continued training, these agents improved markedly relative to EarlyBidder. We also see the performance of the simplest agent, CurrentBid, which does not employ any kind of training, significantly decline relative to the other data-driven agents.

| Agent | Relative Score | | |
|---|---|---|---|
| | Phase I | Phase II | Phase III |
| ATTac-2001$_{ev}$ | $105.2 \pm 49.5$ (2) | $131.6 \pm 47.7$ (2) | $166.2 \pm 20.8$ (1) |
| ATTac-2001$_s$ | $27.8 \pm 42.1$ (3) | $86.1 \pm 44.7$ (3) | $122.3 \pm 19.4$ (2) |
| EarlyBidder | $140.3 \pm 38.6$ (1) | $152.8 \pm 43.4$ (1) | $117.0 \pm 18.0$ (3) |
| SimpleMean$_{ev}$ | $-28.8 \pm 45.1$ (5) | $-53.9 \pm 40.1$ (5) | $-11.5 \pm 21.7$ (4) |
| SimpleMean$_s$ | $-72.0 \pm 47.5$ (7) | $-71.6 \pm 42.8$ (6) | $-44.1 \pm 18.2$ (5) |
| Cond'lMean$_{ev}$ | $8.6 \pm 41.2$ (4) | $3.5 \pm 37.5$ (4) | $-60.1 \pm 19.7$ (6) |
| Cond'lMean$_s$ | $-147.5 \pm 35.6$ (8) | $-91.4 \pm 41.9$ (7) | $-91.1 \pm 17.6$ (7) |
| CurrentBid | $-33.7 \pm 52.4$ (6) | $-157.1 \pm 54.8$ (8) | $-198.8 \pm 26.0$ (8) |

Table 12: The average relative scores ($\pm$ standard deviation) for eight agents in the three phases of our controlled experiment in which the hotel prediction algorithm was varied. The relative score of an agent is its score minus the average score of all agents in that game. The agent's rank within each phase is shown in parentheses.

On the other hand, there are some phenomena in this table that were very surprising to us. Most surprising was the failure of sampling to help. Our strategy relies heavily not only on estimating hotel prices, but also taking samples from the distribution of hotel prices. Yet these results indicate that using expected hotel price, rather than price samples, consistently performs better. We speculate that this may be because an insufficient number of samples are being used (due to computational limitations) so that the numbers derived from these samples have too high a variance. Another possibility is that the method of using samples to estimate scores consistently overestimates the expected score because it assumes the agent can behave with perfect knowledge for each individual sample—a property of our approximation scheme. Finally, as our algorithm uses sampling at several different points (computing hotel expected values, deciding when to buy flights, pricing entertainment tickets, etc.), it is quite possible that sampling is beneficial for some decisions while detrimental for others. For example, when directly comparing versions of the algorithm with sampling used at only subsets of the decision points, the data suggests that sampling for the hotel decisions is most beneficial, while sampling for the flights and entertainment tickets is neutral at best, and possibly detrimental. This result is not surprising given that the sampling approach is motivated primarily by the task of bidding for hotels.

We were also surprised that Cond'lMean$_s$ and Cond'lMean$_{ev}$ eventually performed worse than the less sophisticated SimpleMean$_s$ and SimpleMean$_{ev}$. One possible explanation is that the simpler model happens to give predictions that are just as good as the more complicated model, perhaps because closing time is not terribly informative, or perhaps because the adjustment to price based on current price is more significant. Other things being equal, the simpler model has the advantage that its statistics are based on all of the price data, regardless of closing time, whereas the conditional model makes each prediction based on only an eighth of the data (since there are eight possible closing times, each equally likely).

In addition to agent performance, it is possible to measure the inaccuracy of the eventual predictions, at least for the non-sampling agents. For these agents, we measured the root mean squared error of the predictions made in Phase III. These were: 56.0 for ATTac-2001$_{ev}$, 66.6 for SimpleMean$_{ev}$, 69.8 for CurrentBid and 71.3 for Cond'lMean$_{ev}$. Thus, we see that the lower the error of the predictions (according to this measure), the higher the score (correlation $R = -0.88$).

### 6.3.2 ATTac-2001 vs. EarlyBidder

In a sense, the two agents that finished at the top of the standings in TAC-01 represented opposite ends of a spectrum. The livingagents agent uses a simple open-loop strategy, committing to a set of desired goods right at the beginning of the game, while ATTac-2001 uses a closed-loop, adaptive strategy.

The open-loop strategy relies on the other agents to stabilize the economy and create consistent final prices. In particular, if all eight agents are open loop and place very high bids for the goods they want, many of the prices will skyrocket, evaporating any potential profit. Thus, a set of open-loop agents would tend to get negative scores—the open-loop strategy is a parasite, in a manner of speaking. Table 13 shows the results of running 27 games with 7 copies of the open-loop EarlyBidder and one of ATTac-2001. Although motivated by livingagents, in actuality it is identical to ATTac-2001 except that it uses SimpleMean$_{ev}$ and it places all of its flight and hotel bids immediately after the first flight quotes. It bids only for the hotels that appear in $G^*$ at that time. All hotel bids are for $1001. In the experiments, one copy of ATTac-2001$_s$ is included for comparison. The price predictors are all from Phase I in the preceding experiments. EarlyBidder's high bidding strategy backfires and it ends up overpaying significantly for its goods. As our experiments above indicate, ATTac-2001 may improve even further if it is allowed to train on the games of the on-going experiment as well.

| Agent | Score | Utility |
|---|---|---|
| ATTac-2001 | $2431 \pm 464$ | $8909 \pm 264$ |
| EarlyBidder(7) | $-4880 \pm 337$ | $9870 \pm 34$ |

Table 13: The results of running ATTac-2001 against 7 copies of EarlyBidder over the course of 27 games. EarlyBidder achieves high utility, but overpays significantly, resulting in low scores.

The open-loop strategy has the advantage of buying a minimal set of goods. That is, it never buys more than it can use. On the other hand, it is susceptible to unexpected prices in that it can get stuck paying arbitrarily high prices for the hotel rooms it has decided to buy.

Notice in Table 13 that the average utility of the EarlyBidder's clients is significantly greater than that of ATTac-2001's clients. Thus, the difference in score is accounted for entirely by the cost of the goods. EarlyBidder ends up paying exorbitant prices, while ATTac-2001 generally steers clear of the more expensive hotels. Its clients' utility suffers, but the cost-savings are well worth it.

Compared to the open-loop strategy, ATTac-2001's strategy is relatively stable against itself. Its main drawback is that as it changes its decision about what goods it wants and as it may also buy goods to hedge against possible price changes, it can end up getting stuck paying for some goods that are ultimately useless to any of its clients.

Table 14 shows the results of 7 copies of ATTac-2001 playing against each other and one copy of the EarlyBidder. Again, training is from the seeding round and finals of TAC-01: the agents don't adapt during the experiment. Included in this experiment are three variants of ATTac-2001, each with a different *flight-lookahead* parameter (from the section on "cost of postponing flight commitments"). There were three copies each of the agents with *flight-lookahead* set to 2 and 3 (ATTac-2001(2) and ATTac-2001(3), respectively), and one ATTac-2001 agent with *flight-lookahead* set to 4 (ATTac-2001(4)).

| Agent | Score | Utility |
|-------|-------|---------|
| EarlyBidder | $2869 \pm 69$ | $10079 \pm 55$ |
| ATTac-2001(2) | $2614 \pm 38$ | $9671 \pm 32$ |
| ATTac-2001(3) | $2570 \pm 39$ | $9641 \pm 32$ |
| ATTac-2001(4) | $2494 \pm 68$ | $9613 \pm 55$ |

Table 14: The results of running the EarlyBidder against 7 copies of ATTac-2001 over the course of 197 games. The three different versions of ATTac-2001 had slightly different *flight-lookaheads*.

From the results in Table 14 it is clear that ATTac-2001 does better when committing to its flight purchases later in the game (ATTac-2001(2) as opposed to ATTac-2001(4)). In comparison with Table 13, the economy represented here does significantly better overall. That is, having many copies of ATTac-2001 in the economy does not cause them to suffer. However, in this economy, EarlyBidder is able to invade. It gets a significantly higher utility for its clients and only pays slightly more than the ATTac-2001 agents (as computed by utility minus score).[9]

The results in this section suggest that the variance of the closing prices is the largest determining factor between the effectiveness of the two strategies (assuming nobody else is using the open-loop strategy). We speculate that with large price variances, the closed-loop strategy (ATTac-2001) should do better, but with small price variances, the open-loop strategy could do better.

## 7. Discussion

The open-loop and closed-loop strategies of the previous section differ in their handling of price fluctuation. A fundamental way of taking price fluctuation into account is to place "safe bids." A very high bid exposes an agent to the danger of buying something at a ridiculously high price. If prices are in fact stable then high bids are safe. But if prices fluctuate, then high bids, such as the bids of the stable-price strategy, are risky. In TAC, hotel rooms are sold in a Vickrey-style *n*th price action. There is a separate auction for each day of each hotel and these auctions are done sequentially. Although the order of the auctions is randomized, and not known to the agent, when placing bids in one of these auctions the agent assumes that auction will close next. We assumed in the design of our agent that our bids in one auction do not affect prices in other auctions. This assumption is not strictly true, but in a large economy one expects that the bids of a single individual have a limited effect on prices. Furthermore, the price most affected by a bid is the price of the item being bid on; the effect on other auctions seems less direct and perhaps more limited. Assuming bids in one auction do not affect prices in another, the optimal bidding strategy is the standard strategy for a Vickrey auction—the bid for an item should be equal to its utility to the bidder. So, to place a Vickrey-optimal bid, one must be able to estimate the utility of an item. The utility of owning an item is simply the expected final score assuming one owns the item minus the expected final score assuming one does not own the item. So, the problem of computing a Vickrey-optimal bid can be reduced to the problem of predicting final scores for two alternative game situations. We use two score prediction procedures, which we call the stable-price score predictor (corresponding to Equation 5) and the unstable-price score predictor (Equation 4).

---

9. We suspect that were the agents allowed to retrain over the course of the experiments, ATTac-2001 would end up improving, as we saw in Phase III of the previous set of experiments. Were this to occur, it is possible that EarlyBidder would no longer be able to invade.

**The Stable-Price Score Predictor.** The stable-price score predictor first estimates the expected prices in the rest of the game using whatever information is available in the given game situation. It then computes the value achieved by optimal purchases under the estimated prices. In an economy with stable prices, this estimate will be quite accurate—if we make the optimal purchases for the expected price then, if the prices are near our estimates, our performance will also be near the estimated value.

**The Unstable-Price Score Predictor.** Stable-price score prediction does not take into account the ability of the agent to react to changes in price as the game progresses. Suppose a given room is often cheap but is sometimes expensive. If the agent can first determine the price of the room, and then plan for that price, the agent will do better than guessing the price ahead of time and sticking to the purchases dictated by that price. The unstable price predictor uses a model of the *distribution* of possible prices. It repeatedly samples prices from this distribution, computes the stable-price score prediction under the sampled price, and then takes the average of these stable-price scores over the various price samples. This score prediction algorithm is similar to the algorithm used in Ginsberg's (2001) quite successful computer bridge program where the score is predicted by sampling the possible hands of the opponent and, for each sample, computing the score of optimal play in the case where all players have complete information (double dummy play). While this approach has a simple intuitive motivation, it is clearly imperfect. The unstable-price score predictor assumes both that future decisions are made in the presence of complete price information, and that the agent is free to change existing bids in auctions that have not yet closed. Both of these assumptions are only approximately true at best. Ways of compensating for the imperfections in score prediction were described in Section 5.

**Buy Now or Decide Later.** The trading agent must decide what airline tickets to buy and when to buy them. In deciding whether to buy an airline ticket, the agent can compare the predicted score in the situation where it owns the airline ticket with the predicted score in the situation where it does not own the airline ticket but may buy it later. Airline tickets tend to increase in price, so if the agent knows that a certain ticket is needed it should buy it as soon as possible. But whether or not a given ticket is desirable may depend on the price of hotel rooms, which may become clearer as the game progresses. If airline tickets did not increase in price, as was the case in TAC-00, then they should be bought at the last possible moment (Stone et al., 2001). To determine whether an airline ticket should be bought now or not, one can compare the predicted score in the situation where one has just bought the ticket at its current price with the predicted score in the situation where the price of the ticket is somewhat higher but has not yet been bought. It is interesting to note that if one uses the stable-price score predictor for both of these predictions, and the ticket is purchased in the optimal allocation under the current price estimate, then the predicted score for buying the ticket now will always be higher—increasing the price of the ticket can only reduce the score. However, the unstable-price score predictor can yield an advantage for delaying the purchase. This advantage comes from the fact that buying the ticket may be optimal under some prices but not optimal under others. If the ticket has not yet been bought, then the score will be higher for those sampled prices where the ticket should not be bought. This corresponds to the intuition that in certain cases the purchase should be delayed until more information is available.

Our guiding principle in the design of the agent was, to the greatest extent possible, to have the agent analytically calculate optimal actions. A key component of these calculations is the score predictor, based either on a single estimated assignment of prices or on a model of the probability distribution over assignments of prices. Both score predictors, though clearly imperfect, seem useful. Of these two predictors, only the unstable-price score predictor can be used to quantitatively estimate the value of postponing a decision until more information is available. The accuracy of price estimation is clearly of central importance. Future research will undoubtedly focus on ways of improving both price modeling and score prediction based on price modeling.

## 8. Related and Future Work

Although there has been a good deal of research on auction theory, especially from the perspective of auction mechanisms (Klemperer, 1999), studies of autonomous bidding agents and their interactions are relatively few and recent. TAC is one example. FM97.6 is another auction test-bed, which is based on fishmarket auctions (Rodriguez-Aguilar, Martin, Noriega, Garcia, & Sierra, 1998). Automatic bidding agents have also been created in this domain (Gimenez-Funes, Godo, Rodriguez-Aguiolar, & Garcia-Calves, 1998). There have been a number of studies of agents bidding for a single good in multiple auctions (Ito, Fukuta, Shintani, & Sycara, 2000; Anthony, Hall, Dang, & Jennings, 2001; Preist, Bartolini, & Phillips, 2001).

A notable auction-based competition that was held prior to TAC was the Santa Fe Double Auction Tournament (Rust, Miller, & Palmer, 1992). This auction involved several agents competing in a single continuous double auction similar to the TAC entertainment ticket auctions. As analyzed by Tesauro and Das (2001), this tournament was won by a parasite strategy that, like livingagents as described in Section 6.3, relied on other agents to find a stable price and then took advantage of it to gain an advantage. In that case, the advantage was gained by waiting until the last minute to bid, a strategy commonly known as *sniping*.

TAC-01 was the second iteration of the Trading Agent Competition. The rules of TAC-01 are largely identical to those of TAC-00, with three important exceptions:

1. In TAC-00, flight prices did not tend to increase;

2. In TAC-00, hotel auctions usually all closed at the end of the game;

3. In TAC-00, entertainment tickets were distributed uniformly to all agents

While minor on the surface, the differences significantly enriched the strategic complexity of the game. In TAC-00, most of the designers discovered that a dominant strategy was to defer all serious bidding to the end of the game. A result, the focus was on solving the allocation problem, with most agents using a greedy, heuristic approach. Since the hotel auctions closed at the end of the game, timing issues were also important, with significant advantages going to agents that were able to bid in response to last-second price quotes (Stone & Greenwald, 2003). Nonetheless, many techniques developed in 2000 were relevant to the 2001 competition: the agent strategies put forth in TAC-00 were important precursors to the second year's field, for instance as pointed out in Section 5.1.

Predicting hotel clearing prices was perhaps the most interesting aspect of TAC agent strategies in TAC-01, especially in relation to TAC-00 where the last-minute bidding created essentially a sealed-bid auction. As indicated by our experiments described in Section 6.3, there are many possible approaches to this hotel price estimation problem, and the approach chosen can have a significant impact on the agent's performance. Among those observed in TAC-01 are the following (Wellman, Greenwald, Stone, & Wurman, 2002), associated in some cases with the price-predictor variant in our experiments that was motivated by it.

1. Just use the current price quote $p_t$ (CurrentBid).

2. Adjust based on historic data. For example, if $\Delta_t$ is the average historical difference between clearing price and price at time $t$, then the predicted clearing price is $p_t + \Delta_t$.

3. Predict by fitting a curve to the sequence of ask prices seen in the current game.

4. Predict based on closing price data for that hotel in past games (SimpleMean$_{ev}$, SimpleMean$_s$).

5. Same as above, but condition on hotel closing time, recognizing that the closing sequence will influence the relative prices.

6. Same as above, but condition on full ordering of hotel closings, or which hotels are open or closed at a particular point (Cond'lMean$_{ev}$, Cond'lMean$_s$).

7. Learn a mapping from features of the current game (including current prices) to closing prices based on historic data (ATTac-2001$_s$, ATTac-2001$_{ev}$).

8. Hand-construct rules based on observations about associations between abstract features.

Having demonstrated ATTac-2001's success at bidding in simultaneous auctions for multiple interacting goods in the TAC domain, we extended our approach to apply it to the U.S. Federal Communications Commission (FCC) spectrum auctions domain (Weber, 1997). The FCC holds spectrum auctions to sell radio bandwidth to telecommunications companies. Licenses entitle their owners to use a specified radio spectrum band within a specified geographical area, or *market*. Typically several licenses are auctioned off simultaneously with bidders placing independent bids for each license. The most recent auction brought in over \$16 billion dollars. In a detailed simulation of this domain (Csirik, Littman, Singh, & Stone, 2001), we discovered a novel, successful bidding strategy in this domain that allows the bidders to increase their profits significantly over a reasonable default strategy (Reitsma, Stone, Csirik, & Littman, 2002).

Our ongoing research agenda includes applying our approach to other similar domains. We particularly expect the boosting approach to price prediction and the decision-theoretic reasoning over price distributions to transfer to other domains. Other candidate real-world domains include electricity auctions, supply chains, and perhaps even travel booking on public e-commerce sites.

## Acknowledgements

This work was partially supported by the United States–Israel Binational Science Foundation (BSF), grant number 1999038. Thanks to the TAC team at the University of Michigan for providing the infrastructure and support required to run many of our experiments. Thanks to Ronggang Yu at the University of Texas at Austin for running one of the experiments mentioned in the article. Most of this research was conducted while all of the authors were at AT&T Labs — Research.